\documentclass[letterpaper,conference]{IEEEtran}
\IEEEoverridecommandlockouts
\usepackage{amsmath,epsfig}
\usepackage{amssymb}
\usepackage{amsfonts}
\usepackage{algorithmic}
\usepackage{graphicx}
\usepackage{textcomp}
\usepackage{import}
\usepackage{times}
\usepackage{cleveref}
\usepackage{verbatim}
\usepackage{multirow}
\usepackage{hhline}
\usepackage{threeparttable}
\usepackage{cite}
\usepackage[inline, shortlabels]{enumitem}
\usepackage{xcolor}
\graphicspath{{./}}
\DeclareGraphicsExtensions{.pdf,.jpeg,.png,.eps}
\def\BibTeX{{\rm B\kern-.05em{\sc i\kern-.025em b}\kern-.08em
    T\kern-.1667em\lower.7ex\hbox{E}\kern-.125emX}}

\makeatletter
 \let\old@ps@headings\ps@headings
 \let\old@ps@IEEEtitlepagestyle\ps@IEEEtitlepagestyle
 \def\confheader#1{%
 \def\ps@headings{%
 \old@ps@headings%
 \def\@oddhead{\strut\hfill#1\hfill\strut}%
 \def\@evenhead{\strut\hfill#1\hfill\strut}%
 }%
 \def\ps@IEEEtitlepagestyle{%
 \old@ps@IEEEtitlepagestyle%
 \def\@oddhead{\strut\hfill#1\hfill\strut}%
 \def\@evenhead{\strut\hfill#1\hfill\strut}%
 }%
 \ps@headings%
 }
 \makeatother

\confheader{%
 This paper has been accepted and will be presented at IJCNN 2019
 }
 
\begin{document}

\title{Weakly-Supervised Deep Recurrent Neural Networks for Basic Dance Step Generation \thanks{Research supported by MEXT Grant-in-Aid for Scientific Research (A) 15H01710.}}
%

\author{\IEEEauthorblockN{Nelson Yalta\IEEEauthorrefmark{1},
Shinji Watanabe\IEEEauthorrefmark{2},
Kazuhiro Nakadai\IEEEauthorrefmark{3} and 
Tetsuya Ogata\IEEEauthorrefmark{1}}
\IEEEauthorblockA{\IEEEauthorrefmark{1}\textit{Department of Intermedia Art and Science,
Waseda University, Tokyo, Japan} \\
\IEEEauthorrefmark{2}\textit{Johns Hopkins University, Baltimore, USA} \\
\IEEEauthorrefmark{3}\textit{Honda Research Institute Japan, Saitama, Japan} \\
\IEEEauthorrefmark{1}\textit{nelson.yalta@ruri.waseda.jp}
    } 
}

\maketitle

\begin{abstract}
Synthesizing human's movements such as dancing is a flourishing research field which has several applications in computer graphics.  Recent studies have demonstrated the advantages of deep neural networks (DNNs) for achieving remarkable performance in motion and music tasks with little effort for feature pre-processing. However, applying DNNs for generating dance to a piece of music is nevertheless challenging, because of 
\begin {enumerate*} [1) ]
\item DNNs need to generate large sequences while mapping the music input, 
\item the DNN needs to constraint the motion beat to the music, and 
\item DNNs require a considerable amount of hand-crafted data.  
\end {enumerate*}
In this study, we propose a weakly supervised deep recurrent method for real-time basic dance generation with audio power spectrum as input. The proposed model employs convolutional layers and a multilayered Long Short-Term memory (LSTM) to process the audio input. Then, another deep LSTM layer decodes the target dance sequence. Notably, this end-to-end approach has 
\begin {enumerate*} [1) ]
\item an auto-conditioned decode configuration that reduces accumulation of feedback error of large dance sequence, 
\item uses a contrastive cost function to regulate the mapping between the music and motion beat, and 
\item trains with weak labels generated from the motion beat, reducing the amount of hand-crafted data. 
\end {enumerate*}
We evaluate the proposed network based on 
\begin {enumerate*} [i) ]
\item the similarities between generated and the baseline dancer motion with a cross entropy measure for large dance sequences, and 
\item accurate timing between the music and motion beat with an F-measure. 
\end {enumerate*}
Experimental results revealed that, after training using a small dataset, the model generates basic dance steps with low cross entropy and maintains an F-measure score similar to that of a baseline dancer.

\end{abstract}

\begin{IEEEkeywords}
Deep recurrent networks; Contrastive loss; Dance generation
\end{IEEEkeywords}

\section{Introduction}

%
Dancing is a performing art and expresses meaning or people's emotion \cite{G007}. Dance is composed of sequential rhythmical motion units called basic movements (i.e., basic steps) \cite{G006}. Recently, methods to synthesize dance movements are actively investigated in various domains. Some research has led to the development of applications that go beyond merely generating dance motion for robots \cite{G001,G002}, animated computer graphics, animated choreographies \cite{G003}, and video games \cite{G006}. Besides, various motion and dance generation approaches have been proposed in recent years, including probabilistic models \cite{G003}, Boltzmann machines, and artificial neural networks \cite{G004}. Recent findings demonstrate the remarkable performance of applying deep learning in motion generation \cite{G009,Tang:2018, G021}, music processing \cite{Korzeniowski:17a, G14} and several other tasks. An advantage of using deep learning is the relatively low effort for feature engineering \cite{Meng:18}. Deep learning also enables an end-to-end mapping between the input and output features, reducing the requirement for intermediate hand-crafted annotations \cite{Korzeniowski:17a}. \par
Motivated by the benefits of deep learning, in this study, we explore the application of deep learning models in the generation of basic dance steps. Though deep learning offers the benefits mentioned above, applying it to dance generation can be challenging because of the following three main issues: 
\begin {enumerate*} [1) ]
\item deep learning models may not generate variable-length non-linear sequences \cite{G005} such as dance; 
\item the given model may not be able to constrain the motion beat \cite{G007,G006,G008} to music beat, and 
\item the performance of deep learning models is proportional to the number of training datasets; thus, they require large carefully-labeled datasets for good performance \cite{Meng:18}. 
\end {enumerate*}
\par

\begin{figure*}
  \centering
  \includegraphics[width=\textwidth]{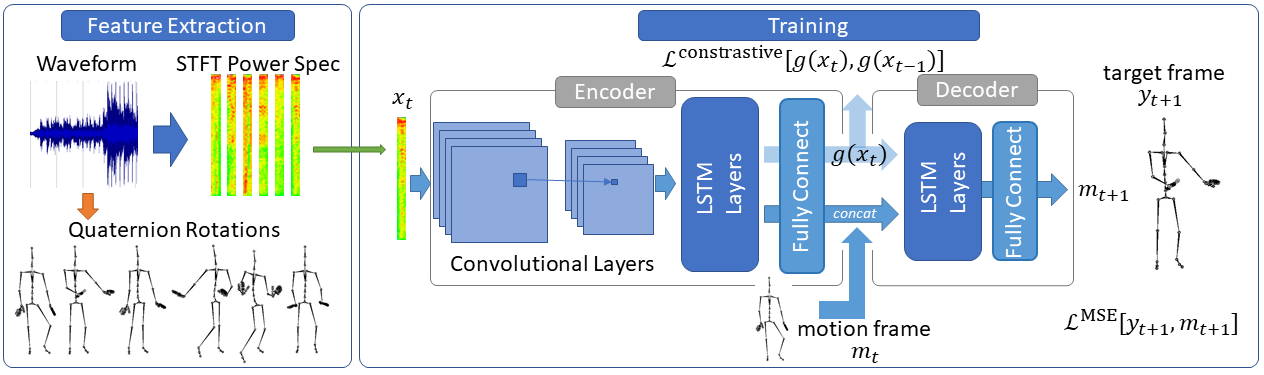}
  \caption{Framework.}
  \label{f:framework}
\end{figure*}

In this study, we propose a weakly-supervised deep recurrent model for generating basic dance synchronized to the music rhythm. Following the research carried out in \cite{Tang:2018} and \cite{G010}, the model proposed in this paper employs multilayered Long Short-Term Memory (LSTM) layers and convolutional layers to encode the audio power spectrum. The convolutional layers reduce the frequency variation of the input audio and the LSTM layers model the time sequence features. Besides, we employ another deep LSTM layer with an  \textbf{auto-conditioned} configuration \cite{G021} to decode the motion. This configuration enables the model to handle a more extended dance sequence with low noise accumulation, which is fed back into the network.
To ensure alignment between the motion and music beat, we utilize a \textbf{contrastive cost function} \cite{G025} for music-motion regulation. The contrastive cost function is a measure of similarities between the given inputs, and it minimizes the distance of the input patterns if the inputs are similar; otherwise, the distance is maximized. Taking advantage of the sequential characteristic of the training, we utilize the Euclidean distance measured by the contrastive loss to track the music beat between two consecutive audio input frames. Then, the model is trained to maximize the distance when the beat is present, i.e., different; otherwise, it minimizes the distance. To constrain the motion beat to the music beat \cite{G007}, we assume that a music beat may be present when a motion beat appears. To avoid additional hand-crafted annotations and reduce the amount of training data, the contrastive cost function employs \textbf{weak labels} (see~\cite{Kumar:2016, Mandel:2008,Meng:18}) that are generated by motion direction. \textit{Weak} labels may yield incomplete, inaccurate or incorrect data \cite{Tian:2013}. However, a contrastive cost function trained with weak labels supports training with a small number of samples and avoids pre-training; therefore, it reduces the need for high computational capabilities. The proposed model demonstrates improved music-motion regulation.\par 
The primary contributions of this study are summarized as follows:
\begin{itemize}
\item We propose a deep recurrent neural network (DRNN) (Section~\ref{ss:drnn}) with an auto-condition configuration  (Section~\ref{ss:autocon}) to generate long dance sequences using the audio spectrum as input. 
\item Using contrastive cost function, we explore the regulation of the alignment between music and motion beat; detecting the music beat between two consecutive audio input frames (Section~\ref{ss:mmc}).
\item Additionally, using motion direction, we explore the generation of \textit{weak} labels for motion-music alignment. Section~\ref{ss:mmc} shows that this configuration reduces the need for additional annotations or hand-crafted labeled data, while we describe the feature extraction and training setup in Section~\ref{s:exp}.
\item We conduct evaluations on the motion beat accuracy and cross entropy of the generated dance relative to the trained music (Section~\ref{s:results}). Furthermore, we demonstrate that the proposed approach increases the precision of the motion beat along with the music beat; moreover, the approach models basic dance steps with lower cross entropy. 
\item Conclusions and suggestions for potential future enhancements of the proposed model are given in Section~\ref{s:conclusion}.
\end{itemize}

\section{Related Works}
\label{s:related}
Several studies have considered different approaches to handle variable-length sequences, such as dance. In \cite{G009}, a factored conditional restricted Boltzmann machine and recurrent neural network (RNN) was employed to tackle the non-linear characteristic of dance. The model maps non-linear characteristics between audio and motion features and generates a new dance sequence. A generative model was presented in \cite{G004} to generate a new dance sequence for a solo dancer. However, the model requires significant computational capabilities or large datasets; moreover, the generated dance is constrained by the trained data. \par
Dancing involves significant changes in motion that occur at regular intervals, i.e., a motion beat (see, e.g.,~\cite{G007,G006,G008}); and when dancing to music, the music and motion beat should be synchronized. In earlier studies, the music beat \cite{Tang:2018} and other musical features \cite{G003} were utilized to improve dance generation. However, the generated features require additional effort to obtain additional annotations. \par
In this paper, we perform dance generation with an audio spectrum as an input and motion-music alignment with weak labels.

\section{Proposed Framework}
\label{s:proposed}
An overview of the proposed system is shown Fig.~\ref{f:framework}.
\subsection{Deep Recurrent Neural Network} 
\label{ss:drnn}
Mapping high-dimensional sequences such as motion is a challenging task for deep neural networks (DNN) \cite{G005} because such sequences are not constrained to a fixed size. Besides, to generate motion from music, the model under consideration must map highly non-linear representations between music and motion \cite{G009}.  In time signal modeling \cite{G15}, DRNNs implemented with LSTM layers displayed remarkable performance and stable training when deeper networks are employed. Furthermore, the application of stacked convolutional layers in a DRNN, referred to as CLDNN, has demonstrated promising results for speech recognition tasks \cite{G010}. In the CLDNN, the convolutional layers reduce the spectral variation of the input sound while the LSTM layers perform time signal modeling. To construct our model, we consider a DRNN with LSTM layers separated into two blocks \cite{G15}: one block plays the role of reducing the music input sequence (encoder), and the other is for the motion output sequence (decoder). This configuration can handle non-fixed dimensional signals, such as motion, and avoids performance degradation due to the long-term dependency of RNNs.\par
\begin{figure}
  \centering
  \includegraphics[width=80mm]{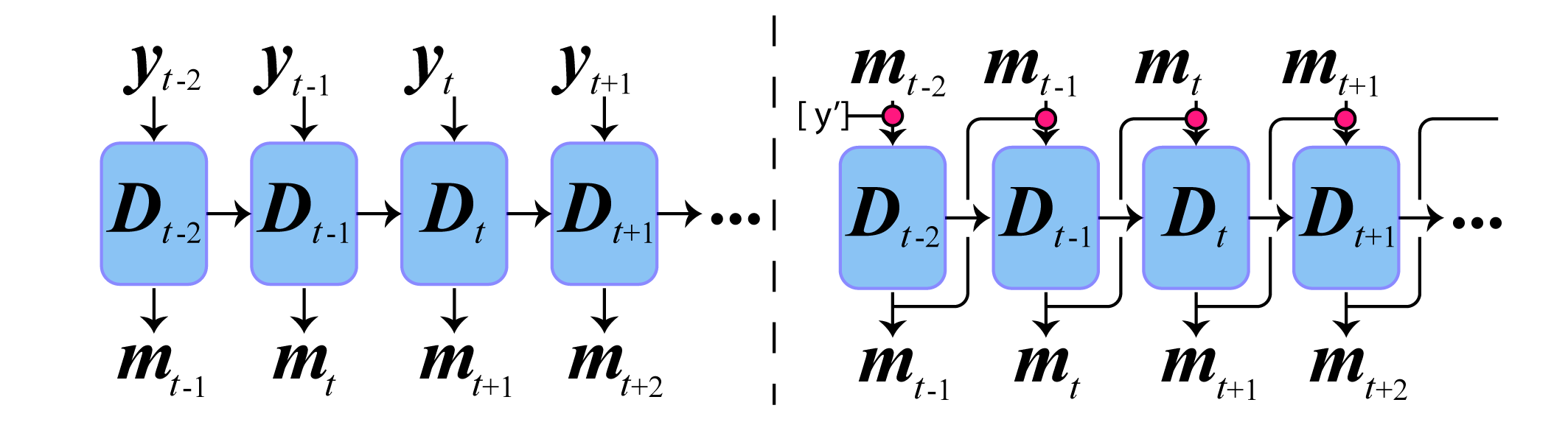}
  \caption{Auto-conditioned decoder.}
  \label{f:autocond}
\end{figure}

The input to the network is the power spectrum from the audio represented as $x_{1:n}= (x_t \in \mathbb{R} ^{b} | t = 1, ..., n)$ with $n$ frames and $b$ frequency bins, and the ground truth sequence is represented as $y_{1:n} = (y(t) \in \mathbb{R} ^{j}| t = 1, ..., n)$ with $n$ frames and $j$ joint axes. 
The following equations show the relations of the motion modeling:
\begin{equation}
g(x_t)=\text{LSTM}^{le}(x'_t); 
\end{equation}
\begin{equation}
m_{t+1} =\text{LSTM}^{ld}(g(x_t)), 
\label{eq:dec1}
\end{equation}
where $g(x_t)$ is the output processed by the encoder with $le$ layers, and $x'_t$ is the output from the convolutional layers. The network output $m_{t+1}$ is processed from the current input and previous states of the decoder with $ld$ layers (Fig.~\ref{f:autocond} left). Then, $m_{t+1}$ is a $l2$-norm model. \par
However, with time-series data (such as dance data), the model may freeze, or the output may diverge from the target due to accumulated feedback errors. To address these issues, the value of the output of the decoder is set to consider auto-regressive noise accumulation by including previously generated step in the motion generation process.

\begin{figure}
  \centering
  \includegraphics[width=80mm]{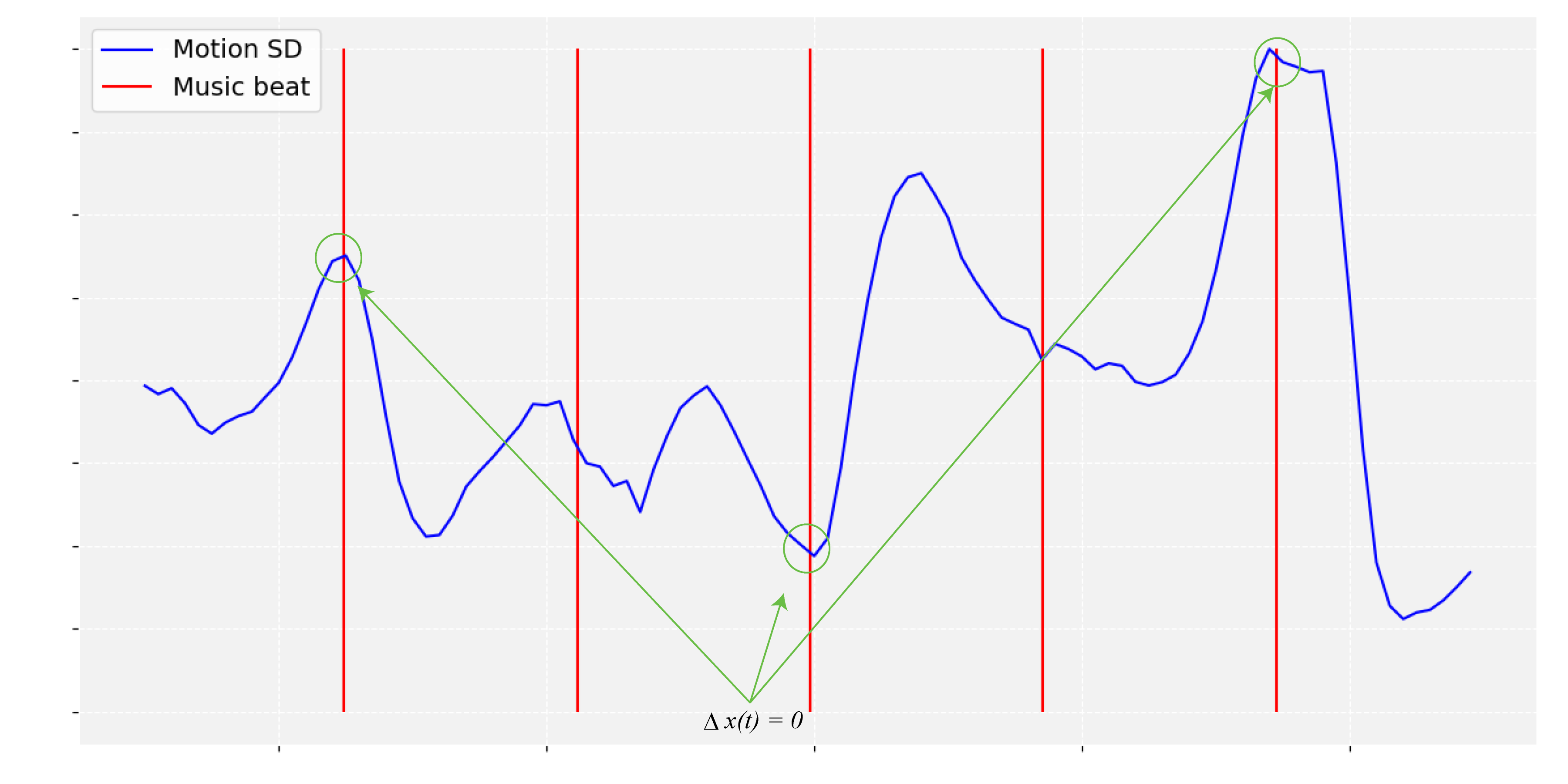}
  \caption{Synchronization of motion and music beat.}
  \label{f:mbmb}
\end{figure}

\subsection{Auto-conditioned Decoder} 
\label{ss:autocon}
A conventional method uses the ground truth of the given sequence as an input to train sequences with RNN models. During evaluations, the model accustomed to the ground truth in the training process may freeze or diverge from the target due to the accumulation of slight differences between the trained and a self-generated sequence.\par
By conditioning the network using its output during training, the auto-conditioned LSTM layer handles errors accumulated during sequence generation. Thus, the network can handle large sequences from a single input, maintain accuracy, and mitigate error accumulation.\par
In the research carried out in \cite{G021} where an auto-conditioned LSTM layer for complex motion synthesis was employed, the conditioned LSTM layer was trained by shifting the input from the generated motion with the ground truth motion after fixed repetitive steps. In the method proposed in this paper, we only employ ground truth motion at the beginning of the training sequence (see Fig.~\ref{f:autocond} right). By modifying Eq.~\ref{eq:dec1}, the generated output is expressed as follows:

\begin{equation}
m_{t+1} =\text{LSTM}^{ld}([g(x_t), y'_t]),
\end{equation}
where 
\begin{equation}
y'_t = 
\begin{cases}
 y_t \text{ if } t = 0, \\
 m_t \text{ otherwise. }
\end{cases}
\end{equation}
The motion error is calculated using a mean squared error (MSE) cost function expressed as follows:

\begin{equation}
\mathcal{L}^{\text{MSE}} = \frac{1}{k}\sum_{i=1}^{k}(y_{t+1}-m_{t+1})^2,
\label{eq:mseloss}
\end{equation}
where $k$ is the training batch size, $y(t+1)$ is the ground truth, and $m(t+1)$ is the generated motion.\par
We employ a zero vector as the input of the first step in our evaluations, followed by a self-generated output to generate the dance until the music stops.\par
\begin{figure}
  \centering
  \includegraphics[width=85mm]{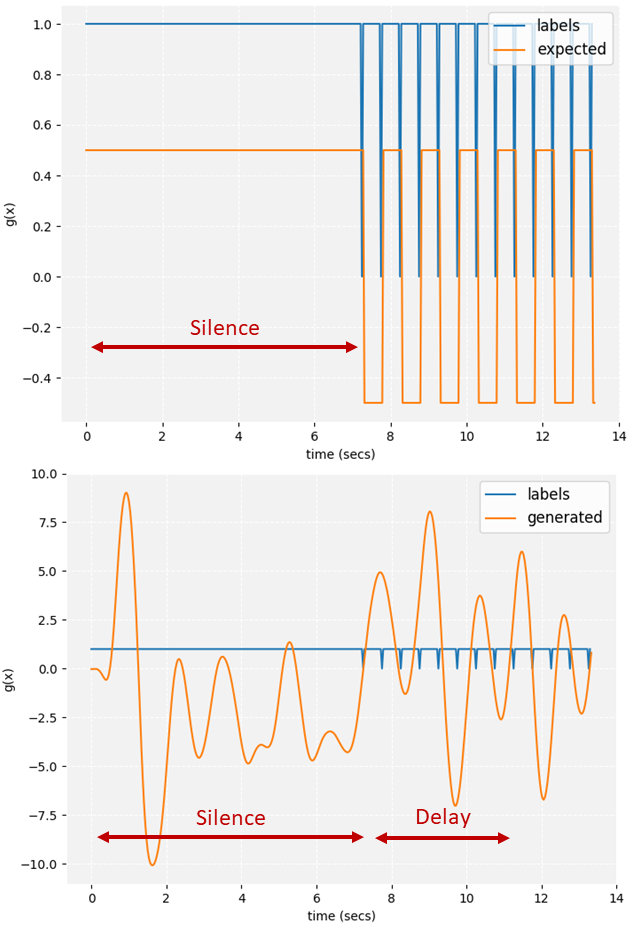}
  \caption{Music-motion regulation. Top: Expected distance, Bottom: Generated distance.}
  \label{f:labels}
\end{figure}

\subsection{Music-motion Alignment Regulation with Weak Labels} 
\label{ss:mmc}
Motion beat is defined as significant changes in movement at regular intervals. An earlier study \cite{G007} revealed that motion beat frames occur when the direction of the movement changes; thus, a motion beat occurs when the speed drops to zero. Furthermore, harmony is a fundamental criterion when dancing to music; hence, the music and motion beat should be synchronized.\par

\begin{figure}
  \centering
  \includegraphics[width=80mm]{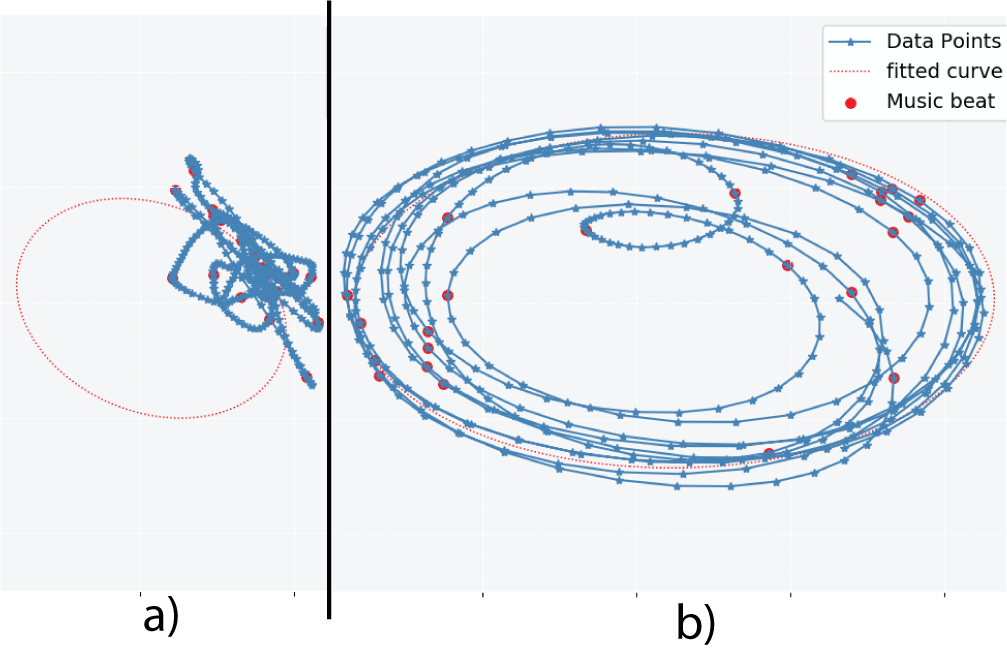}
  \caption{Encoder PCAs: a) DRNN model without contrastive cost function, b) DRNN with contrastive cost function.}
  \label{f:feats}
\end{figure}

For basic dance steps, repetitions of dance steps are given by a repetitive music beat, where the direction of the movement changes drastically (see Fig.~\ref{f:mbmb}). 
To avoid the use of additional information, employing the previous definition we formalize \textit{the extracted music features, which are found to be different compared to the previous frame (i.e., $g(x_t) \neq g(x_{t-1})$) when a beat occurs; otherwise, it may maintain a similar dimension, i.e., $g(x_t) \approx g(x_{t-1})$} (see Fig.~\ref{f:labels}). This procedure generates \textit{weak} labels.\par
Regarding regulation, we employ a contrastive cost function\cite{G025} that maps a similarity metric to the given features. \par
To utilize the contrastive loss, we extract the standard deviation (SD) of the ground truth motion at each frame and compare it to that of the next frame. Then, we assign a label equal to 1 when the motion maintains its direction; otherwise, the label is 0. \par 
 At $t$:
\begin{equation}
dv_t = \text{SD}(y_t)
\end{equation}
\begin{equation}
s_t = \text{sign} (dv_t - dv_{t-1}),
\end{equation}

and at $t+1$:
\begin{equation}
s_{t+1} = \text{sign} (dv_{t+1} - dv_t).
\end{equation}
A label $d$ is expressed as follows:
\begin{equation}
d=\left\{
\begin{array}{ll}
1 & \text{if } s_{t+1}=s_{t}, \\
0 & \text{otherwise.}
\end{array}
\right.
\end{equation}

\begin{table}
\caption{Model architecture.}
\label{t:model}
\centering

\begin{tabular}{cc}
\hline
Layer Name	& Parameters \\\hline\hline
conv1 & $33 \times 2$, 16 channels, stride 1\\
conv2 & $33 \times 2$, 32 channels, stride 1\\
conv3 & $33 \times 2$, 64 channels, stride 1\\
conv4 & $33 \times 2$, 65 channels, stride 1\\
\hline
enc\_lstm1 & 500 units\\
enc\_lstm2 & 500 units\\
enc\_lstm3 & 500 units\\
fc01 & 65-d fc, ELU \\\hline
dec\_lstm1 & 500 units\\
dec\_lstm2 & 500 units\\
dec\_lstm3 & 500 units\\
out & 71-d fc, ELU \\\hline
\hline
\end{tabular}
\end{table}

The contrastive cost function at frame $t+1$ is expressed as:
\begin{equation}
\mathcal{L}^{\text{contrastive}}=\frac{1}{2}(d_{t+1}g^2 + (1-d_{t+1}) \max(1-g,0)^2)
\end{equation}
where $g = ||g(x_{t+1})- g(x_{t})||^2$.\par
Finally, the cost function of the model is formulated as follows:
\begin{multline}
\mathcal{L}_{y}=\mathcal{L}^{\text{MSE}}[y_{t+1},m_{t+1}] +  \max(\mathcal{L}^{\text{contrastive}}[g_{t+1}, g_{t}], 0).
\end{multline}

In this manner, we synchronize the music beat to the motion beat without requiring additional annotations or further information regarding the music beat. Figure~\ref{f:feats} shows the behavior of features from the output encoder after being trained by the contrastive loss. The principal component analysis (PCA) features of the encoder output have a repetitive pattern; moreover, an experimental outcome revealed that they move in an elliptical shape and group the music beats at specific areas.

\subsection{Model Description} 
\label{ss:descript}
The DRNN topology employed in our experiments is comprised of a CLDNN encoder and a deep recurrent decoder (see Table~\ref{t:model}). The CLDNN architecture follows a similar configuration as that considered in \cite{G010}.\par
The input audio features are reduced by four convolutional layers, each followed by a batch normalization \cite{Ioffe:2015A} and an exponential linear unit activation \cite{Clevert:2015}.
Then, three LSTM layers with 500 units each and a fully-connected layer with 65 dimensions complete the structure of the encoder.\par
The input to the decoder consists of the previous motion frame (71 dimensions) and the decoder output (65 dimensions) with a width of 136 dimensions. The decoder is also comprised of three LSTM layers with 500 units each and a fully-connected layer with 71 dimensions. \par
Regarding music-motion control, we add the contrastive cost function after calculating the next step and the MSE cost function.\par

\section{Experiments}
\label{s:exp}
In this study, we conducted a series of experiments to 1) improve motion generation with weakly-supervised learning, and 2) examine the effect of music with different characteristics on the results.\par

\subsection{Data}
Because of a lack of available datasets that include dance synchronized to music, we prepared three datasets only. We restricted the data to small samples using different music genres with different rhythms. \par
\textbf{Hip hop bounce}: This dataset comprises two hip hop music tracks with a repetitive lateral bouncing step to match the rhythm. Each track is three minutes long on the average at 80 to 95 beats per second (bpm).\par
\textbf{Salsa}: This dataset comprises seven salsa tracks (four minutes long on average). All tracks include vocals and rhythms between 95 to 130 bpm. Furthermore, this dataset includes a lateral salsa dance step during instrumental moments and a front-back salsa step during vocal elements.\par
\textbf{Mixed}: This dataset comprises13 music tracks with and without vocal elements (six genres: salsa, bachata, ballad, hip hop, rock, and bossa nova) with an average length of three minutes. Depending on the genre, each track includes up to two dance steps.

\subsection{Audio Feature Extraction}
\label{s:dataprep}
We extracted the power features for each file that was sampled at 16 KHz as follows:
\begin{itemize}
\item First, we utilize white noise to contaminate the audio input at different signal-to-noise ratio (SNR). This allows to augment the number of training files and reduce possible overfitting. The SNR values for the training set were 0 and 10 dB.
\item To synchronize the audio with the motion frame, we extracted a slice of 534 samples (33 milliseconds) of the corresponding position. This extracted slice was converted to \textit{H} short-time Fourier transform (STFT) frames of 160 samples (10 milliseconds) with a shift of 80 samples (5 milliseconds).
\item From the STFT frames, we used the power information, which was normalized between $-0.9 $ and $0.9$ on the \textit{W} frequency bin axis. 
\item Finally, we stacked the \textit{H} frames; thus, the input of the network was a $1 \times W \times H$-dimensional file.
\end{itemize}

\begin{figure}
  \centering
  \includegraphics[width=80mm]{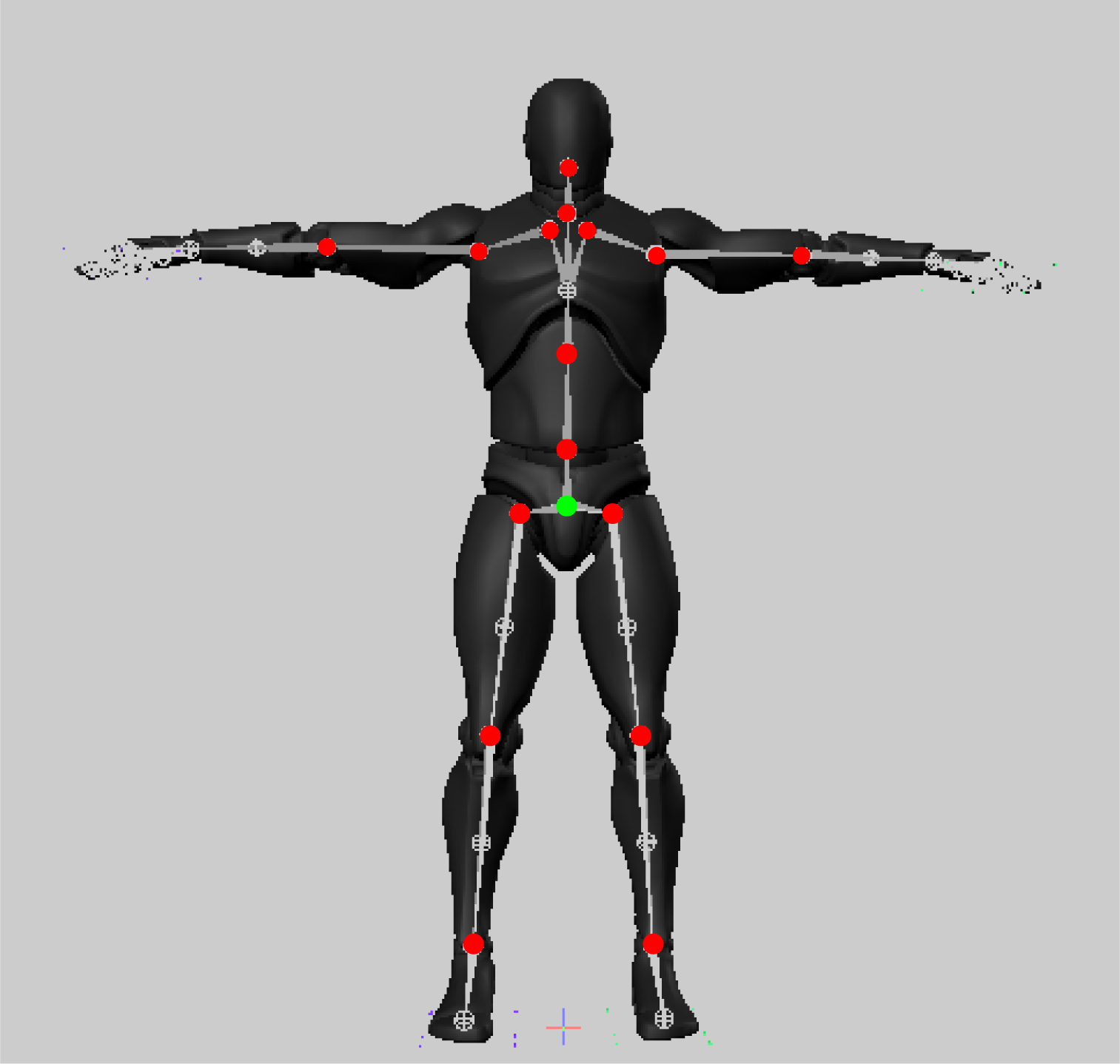}
  \caption{Skeleton of the employed model.}
  \label{f:skeleton}
\end{figure}

\begin{figure*}
  \centering
  \includegraphics[width=\textwidth]{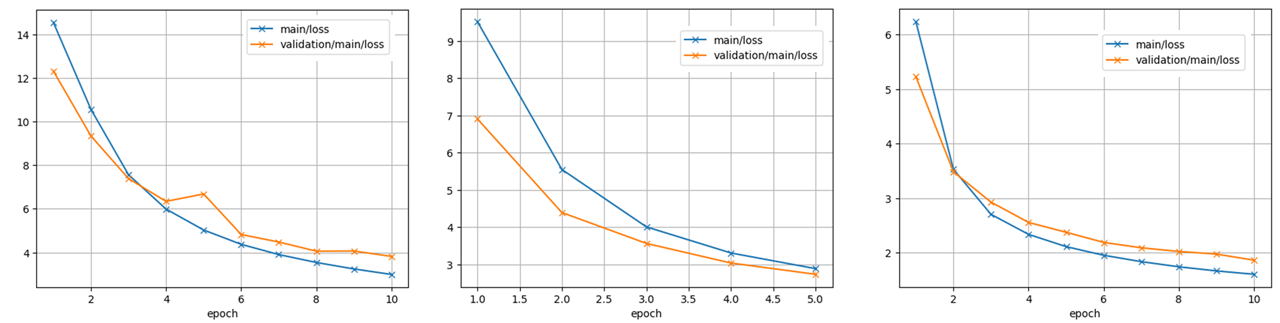}
  \caption{Training loss. Left: Hip hop dataset, Center: Salsa dataset, Right: Mixed genres dataset.}
  \label{f:loss}
\end{figure*}

\begin{table}
\def\arraystretch{1.25}
\caption{Motion features.}
\label{t:motionfeats}
\centering

\begin{tabular}{lcl}
\hline
\textbf{Joint} & \textbf{Type}	& \textbf{Index} \\\hline
Root    &   Translation     & $y^0_t$, $y^1_t$, $y^2_t$ \\
(Root) Pelvis    &   Rotation     & $y^3_t$, $y^4_t$, $y^5_t$, $y^6_t$ \\
Head    &   Rotation     & $y^7_t$, $y^8_t$, $y^9_t$, $y^{10}_t$ \\
Neck    &   Rotation     & $y^{11}_t$, $y^{12}_t$, $y^{13}_t$, $y^{14}_t$ \\
Spine1    &   Rotation     & $y^{15}_t$, $y^{16}_t$, $y^{17}_t$, $y^{18}_t$ \\
Spine2    &   Rotation     & $y^{19}_t$, $y^{20}_t$, $y^{21}_t$, $y^{22}_t$ \\
Left Clavicle    &   Rotation     & $y^{23}_t$, $y^{24}_t$, $y^{25}_t$, $y^{26}_t$ \\
Left Shoulder    &   Rotation     & $y^{27}_t$, $y^{28}_t$, $y^{29}_t$, $y^{30}_t$ \\
Left Forearm    &   Rotation     & $y^{31}_t$, $y^{32}_t$, $y^{33}_t$, $y^{34}_t$ \\
Right Clavicle    &   Rotation     & $y^{35}_t$, $y^{36}_t$, $y^{37}_t$, $y^{38}_t$ \\
Right Shoulder    &   Rotation     & $y^{39}_t$, $y^{40}_t$, $y^{41}_t$, $y^{42}_t$ \\
Right Forearm    &   Rotation     & $y^{43}_t$, $y^{44}_t$, $y^{45}_t$, $y^{46}_t$ \\
Left Thigh    &   Rotation     & $y^{47}_t$, $y^{48}_t$, $y^{49}_t$, $y^{50}_t$ \\
Left Knee    &   Rotation     & $y^{51}_t$, $y^{52}_t$, $y^{53}_t$, $y^{54}_t$ \\
Left Foot    &   Rotation     & $y^{55}_t$, $y^{56}_t$, $y^{57}_t$, $y^{58}_t$ \\
Right Thigh    &   Rotation     & $y^{59}_t$, $y^{60}_t$, $y^{61}_t$, $y^{62}_t$ \\
Right Knee    &   Rotation     & $y^{63}_t$, $y^{64}_t$, $y^{65}_t$, $y^{66}_t$ \\
Right Foot    &   Rotation     & $y^{67}_t$, $y^{68}_t$, $y^{69}_t$, $y^{70}_t$ \\
\hline
\end{tabular}
\end{table}

\subsection{Motion Representation}
For each audio track, we employed the manually selected rotations and root translation (Table~\ref{t:motionfeats}) captured by a single Kinect v2 device at a regular rate of 30 frames per second (see Fig~\ref{f:skeleton}). Then, the captured motion was post-processed and synchronized with the audio data using a motion beat algorithm introduced in \cite{G007}. The motion features are then post-processed from as follows:
\begin{itemize}
\item From the captured data in a hierarchical translation-rotation format, we processed the spatial information (i.e., translation) of the body as a three dimensional vector ($x, y, z$) in meters.
\item Then, the joint rotation in degrees are converted into quaternions \cite{shenitzer:2012} ($q^x_t$, $q^y_t$, $q^z_t$, $q^w_t$) using the tool Transforms3D\footnote{Python Library: https://github.com/matthew-brett/transforms3d} and concatenated them to the translation vector.
\item To avoid the saturation of the activation functions, we normalized each vector component using the maximum value of each component in the range of $-0.9$ to $0.9$. The resultant vector (71 dimensions) was the target of the neural network.
\end{itemize} 
Note that we did not apply any filtering or denoising method to maintain the noisy nature of the motion.  

\subsection{Training procedure}
\label{ss:training}
The models were trained for five to ten epochs using each dataset and the ``chainer'' framework \cite{G024} as an optimization tool. Each training epoch took an average of 60 minutes. \par
The models were trained using an NVIDIA GTX TITAN graphic processing unit. For the optimization, we employed the ``Adam'' solver \cite{G013} with a training mini-batch of 50 files and white noise added to the gradient. Each training batch employed sequences of 150 steps.

\section{Results}
\label{s:results}
\subsection{Training Loss} 
\label{sb:loss}
Figure~\ref{f:loss} shows the training process of the models proposed in this study. The validation data is prepared by contaminating the audio with white noise at an SNR of 5 dB (a different SNR value from those employed for the training set). The models were observed to have fast convergence for all the datasets. However, the salsa dataset required five epochs only while the others required ten epochs each. We assume that the following two characteristics accounts for the short training of the salsa dataset: 1) The salsa dataset has more files than the hip hop bounce dataset which generate more iterations for a single epoch, and 2) the salsa dataset has fewer dance steps than the mixed dataset.

\subsection{Metrics}
\label{sb:metric}
In this study, a quantitative evaluation of the proposed models was performed using the f-score. Using the method employed in \cite{G007}, we extract the motion beat\footnote{Video samples can be found online: https://www.youtube.com/watch?v=VTy0yf-saDQ}  from each dance and obtain the f-score regarding the music beat. To demonstrate the benefits of adding a motion cost function, we compared the performance of the sequence-to-sequence (S2S) and music-motion-loss sequence-to-sequence (S2SMC) models with music beat retrieval frameworks. The models were evaluated under different conditions: clean trained music and white noise at  a music input with a SNR of 20 dB, untrained noises (e.g., claps and crowd noise) at an SNR of 20 dB, clean untrained music tracks of the same genre and music tracks of different genres under clean conditions, and white noise at an SNR of 20 dB. 
Besides, we evaluated the cross entropy of the generated motion and the dancer for the hip hop dataset.

\subsection{Music vs. Motion beat} 
\label{sb:mbvmb}
We compared the results to two music beat retrieval systems; ``madmom'' and the music analysis, retrieval and synthesis for audio signals (MARSYAS) systems. ``Madmom'' \cite{G023} is a Python audio and music signal processing library that employs deep learning to process the music beat, while MARSYAS is an open-source framework that obtains a music beat using an agent-based tempo that processes a continuous audio signal \cite{G022}. The beat annotations of each system were compared to manually annotated ground truth beat data. Furthermore, we compared the ground truth beat data to the motion beat of the dancer for each dance. Here the average f-score was used as a baseline.\par
\begin{table}
\def\arraystretch{1.15}
\caption{F-score and cross entropy of bounce dataset.}
\centering
\resizebox{\columnwidth}{!}{
\begin{tabular}{c|c|c|c|c|c|c|c|c}
\hline
\multirow{4}{*}{Method} & \multicolumn{7}{c|}{F-Score}  & \multirow{4}{*}{Entropy}\\ 
\cline{2-8}
{} & \multicolumn{5}{c|}{Hip Hop}  &  \multicolumn{2}{c|}{Other genres} & {}\\ 
\cline{2-8}

 {} & Clean* & \begin{tabular}{@{}c@{}}White*\\noise\end{tabular}   & Claps & Crowd & Clean & Clean & \begin{tabular}{@{}c@{}}White\\noise\end{tabular}  & {} \\ 
 \hhline{=|=|=|=|=|=|=|=|=}
  \begin{tabular}{@{}c@{}}Madmom\\(Music beat)\end{tabular}   & 89.18 & - & - & - & - & 80.13 & - & -\\  \hline 
  \begin{tabular}{@{}c@{}}Marsyas\\(Music beat)\end{tabular}  & 54.11 & - & - & - & - & 48.89 & - & -\\ 
  \hline 
 \begin{tabular}{@{}c@{}}Dancer\\(baseline)\end{tabular}  & 62.31 & - & - & - & - & 54.49 & - & - \\ \hline 
 S2S & 55.98 & 46.33 & 50.58 & 54.11 & 32.95 & 35.84 & 34.86 & 1.98 \\ 
 S2S-MC & 64.90 & 55.58 & 60.62 & 56.37 & 37.69 & 34.63 & 34.05 & 1.41  \\ \hline
\end{tabular}
}
\begin{tablenotes}\footnotesize
\item[*] *Trained data for S2S and S2SMC
\end{tablenotes}
\label{t:bounce_results}
\end{table}
Table~\ref{t:bounce_results} compares the results of the music beat retrieval frameworks, the motion beat of the dancer (baseline), and the motion beat of the generated dances. It is evident from the evaluation results that the proposed models demonstrate better performance than MARSYAS, and S2SMC outperformed S2S in the evaluations that used clean and noisy data for training. 
It was also found that the addition of different noises to the music input does not critically affect the models when the music input is obtained from the training set. However, the performance is reduced to almost half when an untrained music track of the same or different genres was used as the input. 
Additionally, the proposed models did not outperform a model trained to only process music beat, i.e., MADMOM. We suppose that the music processing approach in our models accounts for this low performance. The proposed model requires three input frames of 33 milliseconds each to generate the motion of a given music beat, while the evaluation only allows a threshold of 70 milliseconds. Furthermore, the accuracy of the training model is reduced due to the delayed reaction of the baseline dancer to the music beat.
S2SMC demonstrated lower cross entropy than S2S, signifying that the dance generated by S2SMC was similar to the trained dance.\par

\begin{table}
\def\arraystretch{1.15}
\caption{F-score of salsa dataset.}
\label{t:salsa_results}
\centering

\begin{tabular}{l|c|c|c|c}
\hline
Method	 & Clean* & White* & Claps & Crowd \\\hline
Madmom & 51.62 & - & - &- \\
Marsyas & 23.38 & - & - &- \\\hline
Dancer (baseline) & 52.82 & - & - &- \\\hline
S2S	&	53.79	& 52.88 & 52.76 & 51.98  \\ 
S2S-MC	& 53.96	& 53.09  & 53.61 & 52.48 \\  \hline
\end{tabular}
\begin{tablenotes}\footnotesize
\item[*]*Trained data for S2S and S2SMC
\end{tablenotes}
\end{table}

Table~\ref{t:salsa_results} shows the f-score for the music tracks trained using the salsa dataset. Both models show better performance than the dancer when tested under the same training conditions, and S2SMC shows better performance than S2S under all conditions. It is worth noting that the size of the dataset influences the performance of the models, besides we employed a larger dataset compared to the that in previous experiment.\par

\begin{table}
\def\arraystretch{1.15}
\caption{F-score of mixed genres.}
\label{t:seq_results}
\centering
\resizebox{\columnwidth}{!}{
\begin{tabular}{l|c|c|c|c|c|c}
\hline
 Method & Bachata* & Ballad* & \begin{tabular}{@{}c@{}}Bossa*\\Nova\end{tabular} & Rock* & Hip Hop* & Salsa* \\ \hline
 Dancer (baseline) & 62.55 & 52.07 & 45.02 & 62.32 & 55.84 & 53.96 \\ \hline 
 S2S  & 60.72 & 49.92 & 46.19 & 60.06 & 64.30   & 52.31 \\ 
 S2S-MC & 56.63 & 48.48 & 40.91 & 64.87 & 63.85 & 53.71 \\ \hline
\end{tabular}
}
\begin{tablenotes}\footnotesize
\item[*]*Trained data for S2S and S2SMC
\end{tablenotes}
\end{table}

Table~\ref{t:seq_results} shows the results of the mixed genre dataset. As can be seen, different results were obtained for each model. The proposed methods do not outperform the baseline, whereas S2S outperformed S2SMC for most genres. The main reason for this difference in the results is the complexity of the dataset and the variety of dance steps relative to the number of music samples; thus, the model could not group the beat correctly.


\section{Conclusion}
\label{s:conclusion}
In this paper, we have presented an optimization technique for weakly-supervised deep recurrent neural networks for dance generation tasks. The proposed model was trained end-to-end and performed better than using only a mean squared cost function. We have demonstrated that the models can generate a correlated motion pattern with a motion beat f-score similar to that of a dancer and lower cross entropy. Besides, the models could be used for real-time tasks because of the low forwarding time (approximately 12 ms). Furthermore, the models show low training time and can be trained from scratch. \par
The proposed model demonstrates reliable performance for motion generation, including music track inputs with a different type of noises. However, the motion pattern is affected by the diversity of the trained patterns and is constrained to the given dataset; this issue will be the focus of future research.

\bibliographystyle{IEEEbib}
\bibliography{refs}

\begin{thebibliography}{10}

\bibitem{G007}
C.~Ho, W.~T. Tsai, K.~S. Lin, and H.~H. Chen,
\newblock ``Extraction and alignment evaluation of motion beats for street
  dance,''
\newblock in {\em 2013 IEEE International Conference on Acoustics, Speech and
  Signal Processing}, May 2013, pp. 2429--2433.

\bibitem{G006}
T.~Kim, S.~I. Park, and S.~Y. Shin,
\newblock ``Rhythmic-motion synthesis based on motion-beat analysis,''
\newblock in {\em ACM SIGGRAPH 2003 Papers}, New York, NY, USA, 2003, SIGGRAPH
  '03, pp. 392--401, ACM.

\bibitem{G001}
J.~L. Oliveira, G.~Ince, K.~Nakamura, K.~Nakadai, H.~G Okuno, and \textit{et
  al.},
\newblock ``Beat tracking for interactive dancing robots,''
\newblock {\em International Journal of Humanoid Robotics}, vol. 12, no. 04,
  pp. 1550023, 2015.

\bibitem{G002}
J.~J. Aucouturier, K.~Ikeuchi, H.~Hirukawa, S.~Nakaoka, and T.~Shiratori
  \textit{et al.},
\newblock ``Cheek to chip: Dancing robots and ai's future,''
\newblock {\em IEEE Intelligent Systems}, vol. 23, no. 2, pp. 74--84, March
  2008.

\bibitem{G003}
S.~Fukayama and M.~Goto,
\newblock ``{Music Content Driven Automated Choreography with Beat-wise Motion
  Connectivity Constraints},''
\newblock in {\em Proceedings of SMC}, 2015.

\bibitem{G004}
L.~Crnkovic{-}Friis and L.~Crnkovic{-}Friis,
\newblock ``Generative choreography using deep learning,''
\newblock {\em CoRR}, vol. abs/1605.06921, 2016.

\bibitem{G009}
O.~Alemi and P.~Pasquier,
\newblock ``Groovenet : Real-time music-driven dance movement generation using
  artificial neural networks,''
\newblock in {\em 23rd ACM SIGKDD Conference on Knowledge Discovery and Data
  Mining.}, 2017.

\bibitem{Tang:2018}
T.~Tang, J.~Jia, and H.~Mao,
\newblock ``Dance with melody: An lstm-autoencoder approach to music-oriented
  dance synthesis,''
\newblock in {\em 2018 {ACM} Multimedia Conference on Multimedia Conference,
  {MM} 2018}, 2018, pp. 1598--1606.

\bibitem{G021}
Z.~Li, Y.~Zhou, S.~Xiao, C.~He, and H.~Li,
\newblock ``Auto-conditioned {LSTM} network for extended complex human motion
  synthesis,''
\newblock {\em CoRR}, vol. abs/1707.05363, 2017.

\bibitem{Korzeniowski:17a}
F.~Korzeniowski and G.~Widmer,
\newblock ``End-to-end musical key estimation using a convolutional neural
  network,''
\newblock {\em CoRR}, vol. abs/1706.02921, 2017.

\bibitem{G14}
V.~Kuleshov, S.~Z. Enam, and S.~Ermon,
\newblock ``Audio super resolution using neural networks,''
\newblock {\em CoRR}, vol. abs/1708.00853, 2017.

\bibitem{Meng:18}
Y.~{Meng}, J.~{Shen}, C.~{Zhang}, and J.~{Han},
\newblock ``{Weakly-Supervised Hierarchical Text Classification},''
\newblock {\em arXiv e-prints}, p. arXiv:1812.11270, Dec. 2018.

\bibitem{G005}
Z.~Gan, C.~Li, R.~Henao, D.~Carlson, and L.~Carin,
\newblock ``{Deep Temporal Sigmoid Belief Networks for Sequence Modeling},''
\newblock {\em Advances in Neural Information Processing Systems}, pp. 1--9,
  2015.

\bibitem{G008}
W.~T. Chu and S.~Y. Tsai,
\newblock ``Rhythm of motion extraction and rhythm-based cross-media alignment
  for dance videos,''
\newblock {\em IEEE Transactions on Multimedia}, vol. 14, no. 1, pp. 129--141,
  Feb 2012.

\bibitem{G010}
T.~N. Sainath, O.~Vinyals, A.~Senior, and H.~Sak,
\newblock ``Convolutional, long short-term memory, fully connected deep neural
  networks,''
\newblock in {\em 2015 IEEE International Conference on Acoustics, Speech and
  Signal Processing (ICASSP)}, April 2015, pp. 4580--4584.

\bibitem{G025}
S.~Chopra, R.~Hadsell, and Y.~LeCun,
\newblock ``Learning a similarity metric discriminatively, with application to
  face verification,''
\newblock in {\em 2005 IEEE Computer Society Conference on Computer Vision and
  Pattern Recognition (CVPR'05)}, June 2005, vol.~1, pp. 539--546 vol. 1.

\bibitem{Kumar:2016}
A.~Kumar and B.~Raj,
\newblock ``Audio event detection using weakly labeled data,''
\newblock in {\em Proceedings of the 24th ACM International Conference on
  Multimedia}, New York, NY, USA, 2016, MM '16, pp. 1038--1047, ACM.

\bibitem{Mandel:2008}
M.~I. Mandel and D.~P.~W. Ellis,
\newblock ``Multiple-instance learning for music information retrieval,''
\newblock in {\em ISMIR}, 2008.

\bibitem{Tian:2013}
F.~Tian and X.~Shen,
\newblock ``Image annotation with weak labels,''
\newblock in {\em Web-Age Information Management}, Jianyong Wang, Hui Xiong,
  Yoshiharu Ishikawa, Jianliang Xu, and Junfeng Zhou, Eds., Berlin, Heidelberg,
  2013, pp. 375--380, Springer Berlin Heidelberg.

\bibitem{G15}
I.~Sutskever, O.~Vinyals, and Q.~V. Le,
\newblock ``Sequence to sequence learning with neural networks,''
\newblock {\em CoRR}, vol. abs/1409.3215, 2014.

\bibitem{Ioffe:2015A}
S.~Ioffe and C.~Szegedy,
\newblock ``Batch normalization: Accelerating deep network training by reducing
  internal covariate shift,''
\newblock in {\em ICML}, 2015.

\bibitem{Clevert:2015}
D.-A. Clevert, T.~Unterthiner, and S.~Hochreiter,
\newblock ``Fast and accurate deep network learning by exponential linear units
  (elus),''
\newblock {\em CoRR}, vol. abs/1511.07289, 2015.

\bibitem{shenitzer:2012}
A.~Shenitzer, B.A. Rosenfeld, and H.~Grant,
\newblock {\em A History of Non-Euclidean Geometry: Evolution of the Concept of
  a Geometric Space},
\newblock Studies in the History of Mathematics and Physical Sciences. Springer
  New York, 2012.

\bibitem{G024}
S.~Tokui, K.~Oono, S.~Hido, and J.~Clayton,
\newblock ``Chainer: a next-generation open source framework for deep
  learning,''
\newblock in {\em Proceedings of Workshop on Machine Learning Systems
  (LearningSys) in The Twenty-ninth Annual Conference on Neural Information
  Processing Systems (NIPS)}, 2015.

\bibitem{G013}
D.~Kingma and J.~Ba,
\newblock ``{Adam: A Method for Stochastic Optimization},''
\newblock {\em arXiv:1412.6980 [cs]}, pp. 1--15, 2014.

\bibitem{G023}
S.~B{\"o}ck, F.~Korzeniowski, J.~Schl{\"u}ter, F.~Krebs, and G.~Widmer,
\newblock ``{madmom: a new Python Audio and Music Signal Processing Library},''
\newblock in {\em Proceedings of the 24th ACM International Conference on
  Multimedia}, Amsterdam, The Netherlands, 10 2016, pp. 1174--1178.

\bibitem{G022}
J.~L. Oliveira,
\newblock ``Ibt: A real-time tempo and beat tracking system,''
\newblock in {\em Proceedings of International Society for Music Information
  Retrieval Conference (ISMIR)}, 2010.

\end{thebibliography}

\end{document}